\documentclass{llncs}%% Use the option review to obtain double line spacing
\usepackage{comment}
\usepackage{wrapfig} % Allows in-line images such as the example fish picture
\usepackage{graphicx,amssymb}
\usepackage{xcolor}
\usepackage{multirow}
\usepackage{subfig}
\usepackage{colortbl}
\usepackage{xcolor}
\begin{document}

\begin{frontmatter}

\title{Deep learning trends for focal brain pathology segmentation in MRI \thanks{This is a pre-print version. The original paper is published in {\it Machine Learning for Health Informatics}.}}

\author{Mohammad Havaei\thanks{\texttt{seyed.mohammad.havaei@usherbrooke.ca}} \inst{1} 
 \and Nicolas Guizard\inst{2} \and Hugo Larochelle \inst{1} \inst{4}\\ \and 
Pierre-Marc Jodoin \inst{1}\inst{3}  
}
\institute{Universit\'e de Sherbrooke, Canada\\
\and
Imagia cybernetics, Canada 
\and Imeka Inc., Canada
\and Google, Canada
}

\titlerunning{Deep learning in brain pathology segmentation}
\authorrunning{Havaei et al.}

\maketitle

\begin{abstract}
Segmentation of focal (localized) brain pathologies such as brain tumors and brain lesions caused by multiple sclerosis and ischemic strokes are necessary for medical diagnosis, surgical planning and disease development as well as other applications such as tractography. 
Over the years, attempts have been made to automate this process for both clinical and research reasons. In this regard, machine learning methods have long been a focus of attention. Over the past two years, the medical imaging field has seen a rise in the use of a particular branch of machine learning commonly known as deep learning. 
In the non-medical computer vision world, deep learning based methods have obtained state-of-the-art results on many datasets.  Recent studies in computer aided diagnostics have shown deep learning methods (and especially convolutional neural networks - CNN) to yield promising results. In this chapter, we provide a survey of CNN methods applied to medical imaging with a focus on brain pathology segmentation. In particular, we discuss their characteristic peculiarities and their specific configuration and adjustments that are best suited to segment medical images. We also underline the intrinsic differences deep learning methods have with other machine learning methods.

\end{abstract}


\end{frontmatter}

%% \linenumbers

%% main text

\section{Introduction}

    Focal pathology detection of the central nerveous system (CNS), such as lesion, tumor and hemorrhage is primordial for accurate diagnosis, treatment and for future prognosis. The location of this focal pathology in the CNS, determines the related symptoms but clinical examination might not be sufficient to clearly identify the underlying pathology.  
    Ultrasound, computer tomography and conventional MRI acquisition protocols are standard image modalities used clinically. The qualitative MRI modalities T1 weighted (T1), T2 weighted (T2), Proton density weighted (PDW), T2-weighted FLAIR (FLAIR) and contrast-enhanced T1 (T1C), diffusion weighted MRI and functional MRI are sensitive to the inflammatory and demyelinating changes directly associated with the underlying pathology. As such, MRI is often used to detect, monitor, identify and quantify the progression of the diseases.  
    
    For instance, in multiple sclerosis (MS), T2 lesions are mainly visible in white matter (WM), but can be found also in gray matter (GM). MS lesions are more frequently located in the peri-ventricular or sub-cortical region of the brain. They vary in size, location and volume, but are usually elongated along small vessels. These lesions are highly heterogeneous and include different underlying processes: focal breakdown of the blood-brain barrier, inflammation, destruction of the myelin sheath (demyelination), astrocytic gliosis, partial preservation of axons and remyelination. Similarly, in Alzheimer's disease (AD), white matter hyperintensity (WMH), which are presumed to be from  vascular origin, are also visible in FLAIR images and are believed to be a biomarker of the disease. Similar to vascular hemorrhages, ischemic arterial or venous strokes can be detected with MRI. MRI is also used for brain tumor segmentation which is necessary for monitoring the tumor growth or shrinkage, for tumor volume measurement and also for surgical and radiotherapy planning. 
For glioblastoma segmentation, different MRI modalities highlight different tumor sub-regions . For example, T1 is the most commonly used modality for structural analysis and distinguishing healthy tissues. In T1C, the borders of the glioblastoma are enhanced. This modality is most useful for distinguishing the active part of the glioblastoma from the necrotic parts. In T2, the edema region appears bright and using FLAIR, we can distinguish between the edema and CSF. This is  possible because 
CSF appears dark in FLAIR. 

 The sub-regions of a glioblastoma are as follows: 

\begin{itemize}
\item\textit{Necrosis}\textendash The dead part of the tumor. 
\item\textit{Edema}\textendash The swelling caused by the tumor. As the tumor grows, it can block the cerebrospinal fluid from going out of the brain. New blood vessels growing in and near the tumor can also lead to swelling.   
\item\textit{Active-enhanced}\textendash Refers to the part of the tumor which is enhanced in T1C modality. 
\item\textit{Non-enhanced}\textendash Refers to the part of the tumor which is not enhanced in T1C modality. 
\end{itemize}

%\paragraph{Challenges with brain tumor segmentation}
There are many challenges associated with the segmentation of a brain pathology. The main challenges come from the data acquisition procedure (MRI in our case) as well as from the nature of the pathology.  Those challenges can be summarized as follows:
\begin{itemize}
\item  Certainly, the most glaring issue with MR images comes from the non-standard intensity range obtained from different scanners.  Either because of the various magnet strengths (typically 1.5, 3 or 7 Tesla) or because of different acquisition protocols, the intensity values of a brain MRI, is often very different from one hospital to another, even for the same patient.

\item There are no reliable shape or intensity priors for brain tumors/lesions. Brain pathology can appear anywhere in the brain, they can have any shape (often with fuzzy borders) and come with a wide range of intensities.  Furthermore, the intensity range of such pathology may overlap with that of healthy tissue making computer aided diagnosis (CAD) complicated.

\item MR images come with a non negligible amount of white Rician noise introduced during the acquisition procedure. 

\item Homogeneous tissues (typically the gray and the white matter) often suffer from spatial intensity variations along each dimension. This is caused by a so-called bias field effect.  The MRI bias is a smooth low-frequency signal that affects the image intensities.  This problem calls for a bias field correction pre-processing step which typically increases intensity values at the periphery of the brain. 

\item MR images may have  non-isotopic resolution, leading to low resolution images, typically along the coronal and the saggital views. 

\item The presence of a large tumor or lesion in the brain, may warp the overall structure of the brain, thus making some procedures impossible to perform.  For example, large tumors may affect the overall symmetry of the brain, making left-right symmetry features impossible to compute. %\cite{SOMETHING}.  
Also, brains with large tumors can hardly be registered onto a healthy brain template.

\end{itemize}

Methods relying on machine learning also have their own challenges when processing brain images. To count a few:

\begin{itemize}
    \item Supervised  methods require a lot of labeled data in order to generalize well to unseen examples. As opposed to non-medical computer vision applications, acquiring medical data is time consuming, often expensive and requires the non-trivial approval of an ethical committee as well as the collaboration of non-research affiliated staff.  Furthermore, the accurate ground truth labeling of 3D MR images is time consuming and expensive, as it has to be done by highly trained personnel (typically neurologists).  As such, publicly-available medical datasets are rare and often made of a limited number of subjects.  One consequence of not having enough labeled data is that the models trained on such datasets are prone to overfitting and perform poorly on new subjects.
    \item In supervised learning, we typically estimate by maximum likelihood and thus assume that the examples are identically distributed.  Unfortunately, the intensity variation from one MRI machine to another, often violates that assumption.
    Large variations in the data distribution  can be leveraged by having a sufficiently large training dataset, which is almost never the case with medical images.
    \item Classic machine learning methods rely on computing high dimensional feature vectors, which can make them computationally inefficient both memory-wise and processing-wise.
    \item Generally in brain tumor/lesion segmentation, ground truth is heavily imbalanced since regions of interest are very small compared to the whole brain. This is very unfortunate for many machine learning methods such as neural networks which work best when classes have similar size. 
    \item Because of the variability of the data, there is no standard pre-processing procedure.
\end{itemize}

 Most brain lesion segmentation methods use hand-designed features~\cite{braintumorsegmentationdotorg,Menze2014}.  These methods implement a classical machine learning pipeline according to which features are first extracted and then given to a classifier whose training procedure does not affect the nature of those features.

An alternative would be to {\it learn} such a hierarchy of increasingly complicated features (i.e.\ low, mid and high level features). Deep neural networks (DNNs) have been shown to be successful in learning task-specific feature hierarchies \cite{bengio2013}. Importantly, a key advantage of DNNs is that they allow to learn MRI brain-pathology-specific features that combine information from across different MRI modalities. Also,  convolutions are very efficient and can make predictions very fast. We investigate several choices for training Convolutional Neural Networks (CNNs) for this problem and report on their advantages, disadvantages and performance.
  Although CNNs first appeared over two decades ago~\cite{lecun1998},
they have recently become a mainstay for the computer vision community due to their record-shattering performance in the ImageNet Large-Scale Visual Recognition Challenge~\cite{Krizhevsky-2012-small}.
%as well as the well-known CIFAR-10~\citep{Goodfellow_maxout_2013} evaluation. % DWF: CIFAR10 is not nearly as big a deal to CV people.
While CNNs have also been successfully applied to segmentation problems~\cite{Alvarez2012,long2015fully,SimulDetectSegm}, most of the previous work have focused on non-medical tasks and many involve architectures that are not well suited to medical imagery or brain tumor segmentation in particular. 

Over the past two years, we have seen an increasing use of deep learning in health care and more specifically in medical imaging segmentation. This increase can be seen in recent Brain Tumor Segmentation challenges (BRATS) which is held in conjunction with Medical Image Computing and Computer Assisted Intervention (MICCAI). While in 2012 and 2013 none of the competing methods used DNNs, in 2014, 2 of the 15 methods and in 2015, 7 of the 13 methods taking part in the challenge were using DNNs. In this work, we explore a number of approaches based on deep neural network architectures applied to brain pathology segmentation. 

\section{Glossary}
    \paragraph{\bf Cerebral spinal fluid (CSF)}: a clear, colorless liquid located in the middle of the brain. 
    \paragraph{\bf Central nervous system (CNS)}: part of the nervous system consisting of the brain and the spinal cord.
    \paragraph{\bf Diffusion weighted image (DWI)}: MR imaging technique, measuring the diffusion of water molecules within tissue voxels.   DWI is often used to visualize hyperintensities.  
    \paragraph{\bf Deep Neural Network (DNN)}:  an artificial intelligence system inspired from human nervous system, where through a hierarchy of layers, the model learns a hierarchy of low to high end features.
    \paragraph{\bf Convolutional Neural Network (CNN)}: a type of DNN adopted for imagery input. The number of parameters in a CNN is significantly less than that of a DNN due to a parameter sharing architecture made feasible by convolutional operations.%As opposed to DNNs where a hidden layer is connected to all elements in the input, a hidden CNN unit is connected to a subset of the input.
    \paragraph{\bf FLAIR image}: an MRI pulse sequence that suppresses fluid (mainly cerebrospinal fluid (CSF)) while enhancing edema.
    \paragraph{\bf Gray matter (GM)}: a large region located on the surface of the brain consisting mainly of nerve cell bodies and branching dendrites.
    \paragraph{\bf High-grade glioma}:  malignant brain tumors of types 3 and 4.
    \paragraph{\bf Low-grade glioma}: slow growing brain tumors of types 1 and 2. 
    \paragraph{\bf Multiple sclerosis (MS)}: a disease of the central nervous system attacking the myelin, the insulating sheath surrounding the nerves.
    \paragraph{\bf Overfitting}: in machine learning the {\it overfitting} phenomenon occurs when the model is too complex relative to the number of observations. Overfitting reduces the ability of the model to generalize to unseen examples.     
    \paragraph{\bf Proton density weighted (PDW) image}: an MR image sequence used to measure the density of protons; an intermediate sequence sharing some features of both T1 and T2. In current practices, PDW is mostly replaced by FLAIR.
    \paragraph{\bf T1-weighted image}: one of the basic MRI pulse sequences showing the difference in the T1 relaxation times of tissues \cite{t1_url}.
    \paragraph{\bf T1 Contrast-enhanced image}: a T1 sequence, acquired after a gadolinium injection. Gadolinium changes the signal intensities by shortening the T1 time in its surroundings. Blood vessels and pathologies with high vascularity appear bright in T1 weighted post gadolinium images.
    \paragraph{\bf T2-weighted image}: one of the basic MRI pulse sequences. The sequence highlights differences in the T2 relaxation time of various tissue\cite{t2_url}.
    \paragraph{\bf White matter hyperintensity}:  changes in the cerebral white matter in aged individuals or patients suffering from a brain pathology~\cite{whitematterhyperintensity}.

\section{Datasets}
In this section, we describe some of the most widely-used public datasets for brain tumor/lesion segmentation. 

\paragraph{{\bf BRATS benchmark}}
The Multimodal BRain Tumor image Segmentation (BRATS), is a challenge held annually in conjunction with the MICCAI conference since 2012. The BRATS 2012 training data consist of 10 low- and 20 high-grade glioma MR images whose voxels have been manually segmented with three labels ({\it healthy}, {\it edema} and {\it core}). The challenge data consist of 11 high- and 5 low-grade glioma subjects and no ground truth is provided for this dataset. Having only two basic tumor classes is insufficient due to the fact that the {\it core} label contains structures which vary in different modalities. For this reason, the BRATS 2013 dataset contains the same training data but was manually labeled into 5 classes; {\it healthy}, {\it necrosis}, {\it edema} {\it non-enhanced} and {\it enhanced tumor}. There are also two test sets available for BRATS 2013 which do not come with ground truth; the {\it leaderboard} dataset which contains the BRATS 2012 challenge dataset with additional 10 high-grade glioma patients and the BRATS 2013 {\it challenge} dataset which contains 10 high-grade glioma patients. 
The above mentioned datasets are available for download through the challenge website \cite{VSD}.

For BRATS 2015, the size of the dataset was increased extensively\footnote{Note that the BRATS organizers released a dataset in 2014 which was later removed from the web. This version of the dataset is no longer available.}.  BRATS 2015 contains 220 subjects with high-grade and 54 subjects with low grade gliomas for training and 53 subjects with mixed high and low grade gliomas for testing. Similar to BRATS 2013, each brain from the training data, comes with a 5 class segmentation ground truth. BRATS 2015 also contains the training data of BRATS 2013. %The ground truth for the rest of the training subjects are generated by a voted average of the segmented results of the top performing methods in BRATS 2013 and BRATS 2012.  
The ground truth for the rest of the training subjects are generated automatically with the integration of the top performing methods in BRATS 2013 and BRATS 2012. 
Although some of the automatically generated ground truths have been refined manually by a user, some challenge participants have decided to remove subjects with heavily corrupted ground truths from their training data~\cite{havaeic,urban2014,Kleesiek2014}. This dataset can be downloaded through the challenge website \cite{VSD}. 

All BRATS datasets, share four MRI modalities namely; T1, T1C, T2, FLAIR. Image modalities for each subject are co-registered to T1C. Also, all images are skull stripped.

Quantitative evaluation of the model's performance on the test set is achieved by uploading the segmentation results to the online BRATS evaluation system~\cite{VSD}. The online system provides the quantitative results as follows: 
The tumor structures are grouped in 3 different tumor regions. This is mainly due to practical clinical applications. 
As described by Menze et al. (2014)~\cite{Menze2014}, tumor regions are defined as:

\begin{enumerate}

{\setlength\itemindent{25pt} \item The {\it complete} tumor region (including all four tumor structures).}

{\setlength\itemindent{25pt}\item The {\it core} tumor region (including all tumor structures exept ``edema").}

{\setlength\itemindent{25pt}\item The {\it enhancing} tumor region (including the ``enhanced tumor" structure).}

\end{enumerate} 

%Depending on the year the challenge was held, different evaluation metrics have been considered.  
For each tumor region, Dice, Sensitivity, Specificity, Kappa as well as the Hausdorff distance are reported.  
The online evaluation system provides a ranking for every method submitted for evaluation. This includes methods from the 2013 BRATS challenge published in \cite{Menze2014} as well as anonymized unpublished methods for which no reference is available.

%\begin{eqnarray}
%Dice(P,T) &=& \frac{|P_1 \wedge T_1|}{(|P_1|+|T_1|)/2}, \nonumber \\
%Sensitivity(P,T) &=& \frac{|P_1 \wedge T_1|}{|T_1|}, \nonumber \\
%Specificity(P,T) &=& \frac{|P_0 \wedge T_0|}{|T_0|}, \nonumber
%\end{eqnarray}

%where $P$ represents the model predictions and $T$ represents the ground truth labels. We also note as $T_1$ and $T_0$ the subset of voxels predicted as positives and negatives for the tumor region in question. Similarly for $P_1$ and $P_0$.  

\paragraph{{\bf ISLES benchmark}}
 Ischemic Stroke Lesion Segmentation (ISLES) challenge started in 2015 and is held in conjunction with the Brain Lesion workshop as part of MICCAI. ISLES has two categories with individual datasets; sub-acute ischemic stroke lesion segmentation (SISS) and acute stroke outcome/penumbra estimation (SPES) datasets~\cite{isles2015}. Similar to BRATS, an online evaluation system is available to evaluate the segmentation outputs of the test subjects.
 
\paragraph{SISS} contains 28 subjects with four modalities, namely: FLAIR, DWI, T2 TSE (Turbo Spin Echo), and T1 TFE (Turbo Field Echo). The challenge dataset consists of 36 subjects.  The evaluation measures used for the ranking are the Dice coefficients, the average symmetric surface distance, and the Hausdorff distance.
\paragraph{SPES} dataset contains 30 subjects with 7 modalities namely:
CBF (Cerebral blood flow), CBV (cerebral blood volume), DWI, T1C, T2, Tmax and TTP (time to peak). The challenge dataset contains 20 subjects. 
Both datasets provide pixel level ground truth of the abnormal areas (2 class segmentation). The metrics used to gauge performances are the Dice score, the Hausdorff distance, the recall and precision as well as the average symmetric surface distance (ASSD).

\paragraph{\bf MSGC benchmark}
The MSGC dataset which was introduced at MICCAI 2008 \cite{styner2008MSGC}, provides 20 training MR cases with manual ground truth MS lesion segmentation and 23 testing cases from the Boston Children’s Hospital (CHB) and the University of North Carolina (UNC). For each subject, T1, T2 and FLAIR are provided which are co-registered. While lesions masks for the 23 testing cases are not available for download, an automated system is available to evaluate the output of a given segmentation algorithm. The MSGC benchmark provides different metric results normalized between 0 and 100, where 100 is a perfect score and 90 is the typical score of an independent rater \cite{styner2008MSGC}. The different metrics (volume difference "VolD", surface distance "SurfD", true positive rate "TPR" and false positive rate "FPR") are measured by comparing the model output segmentation to the manual segmentation of two experts  at CHB and UNC. 

\section{State-of-the-art}
In this section, we present a brief overview of some methods used to segment brain lesions and brain tumors from MR images.

\subsection{Pre deep learning era}
These methods can be grouped in two major categories: {\it semi-automatic} and {\it automatic} methods. Semi-automatic (or interactive) methods are those relying on user intervention. Many of these methods rely on active deformable models ({\it e.g.} snakes) where the user initializes the tumor contour  \cite{Jiang2004,Wang2009}.  Other semi-automatic methods use classification  which the input to the model is given through regions of interest drawn from inside and outside of the tumor~\cite{kaus1999,zhang2004,havaei2014,havaeia,bauer2013}. Semi-automatic methods are appealing in medical imaging applications since the datasets are generally very small~\cite{holzinger2016interactive,girardiinteractive}.
Automatic methods on the other hand are those for which no user interaction is made. These methods can be divided into two groups;
The first group of methods are based on {\it anomaly} detection, where the model estimates intensity similarities between the query subject and an atlas. By doing so, brain regions which deviate from healthy tissue are detected. These techniques have shown good results in structural segmentation when using non-linear registration~\cite{guizard2015rotation,rexilius2007,prastawa2003b,khotanlou2007}.

The second group of methods are {\it machine learning methods}, where a discriminative model is trained using {\it pre-defined} features of the input modalities. After integrating different intensity and texture features, a classifier is trained to decide to which class each voxel belongs to. Random forests have been particularly popular. Reza et al.~\cite{Reza2013} used a mixture of intensity and texture features to train a random forest for voxelwise classification. One problem with this approach is that the model should be trained in a high-dimensional feature space. For example, Festa et al.~\cite{festa2013} used a feature space of 300 dimensions and the trained random forest comprised of 50 trees.  To train more descriptive classifiers, some methods have taken the approach of adding classes to the ground truth \cite{bauer2012,zhao2012}. Tustison et al.~\cite{tustison2015} does this by  using Gaussian Mixture Models (GMMs) to get voxelwise tissue probabilities for WM, GM, CSF, edema, non-enhancing tumor, enhancing tumor, necrosis. The GMM is initialized with prior cluster centers learnt from the training data. The voxelwise probabilities are used as input features to a random forest. The intuition behind increasing the number of classes is that the distribution of the healthy class is likely to have different modes for WM, GM and the CSF and so the classifier would be more confidant if it tries to classify them as separate classes. 
Markov random fields (MRF) as well as conditional random fields (CRF) are sometime used to regularize the predictions \cite{meier2013,Havaei2016,Lee05,tustison2015}. Usually, the pairwise weights in these models are either fixed \cite{Havaei2016} or determined by the input data. They work best in the case of weak classifiers such as k-nearest neighbor (kNN) or decision trees and become less beneficial when using stronger classifiers such as convolutional neural networks \cite{roth2015}. 

Deformable models can also be used as post-processing, where an automatic method is used to initialize the counter as opposed to user interaction in semi-automatic methods \cite{Ho2002,rexilius2007,prastawa2003b,khotanlou2007}. 

\subsection{Deep learning based methods} 
\label{deep-learning-era}

As mentioned before, classical machine learning methods in both automatic and semi-automatic approaches use pre-defined (or hand-crafted) features which might or might not be useful in the training objective. Oppose to that, deep learning methods {\it learn} features specific to the task at hand. Moreover, these features are learnt in a hierarchy of increasing feature complexity, which results in more robust features.

Recently, deep neural networks have proven to be very promising for medical image segmentation. In the past two years, we have seen an increase in use of neural networks applied to brain tumor and lesion segmentations. Notable mentions are the MICCAI brain tumor segmentation challenges (BRATS) in 2014 and 2015 and the ISLES challenge in 2015 where the top performing methods were taking use of convolutional neural networks \cite{braintumorsegmentationdotorg,braintumorsegmentationdotorg15}. 

 In spite of the fact that CNNs were originally developed for image classification, it is possible to use them in a segmentation framework. A simple approach is to train the model in a {\it patch wise} fashion as in \cite{ciresan2012deep}, where for every training (or testing) pixel $i$, a patch $\textbf{x}_i$ of size $n \times n$ around $i$ is extracted, and the goal is to identify class label of the center pixel.

Although MRI segmentation is a 3D problem, most methods take a 2D approach by processing the MRI slice by slice. For these methods, training is mostly done patch wise on the axial slices. Zikic et al.~\cite{zikic2014} use a 3 layer model with 2 convolutional layers and one dense layer. The input size of the model is $19 \times 19$, however, since the inputs have been downsampled by a factor of 2, the effective receptive field size is $38 \times 38$. {\it Max pooling} with a stride of 3 is used at the first convolutional layer. During test time, downsampled patches of $19 \times 19$ are presented to the model in sliding window fashion to cover the entire MRI volume. The resulting segmentation map is upsampled by a factor of two in order to have the same size as the input.

\begin{figure}[t]
\centering
\includegraphics[width= \linewidth]{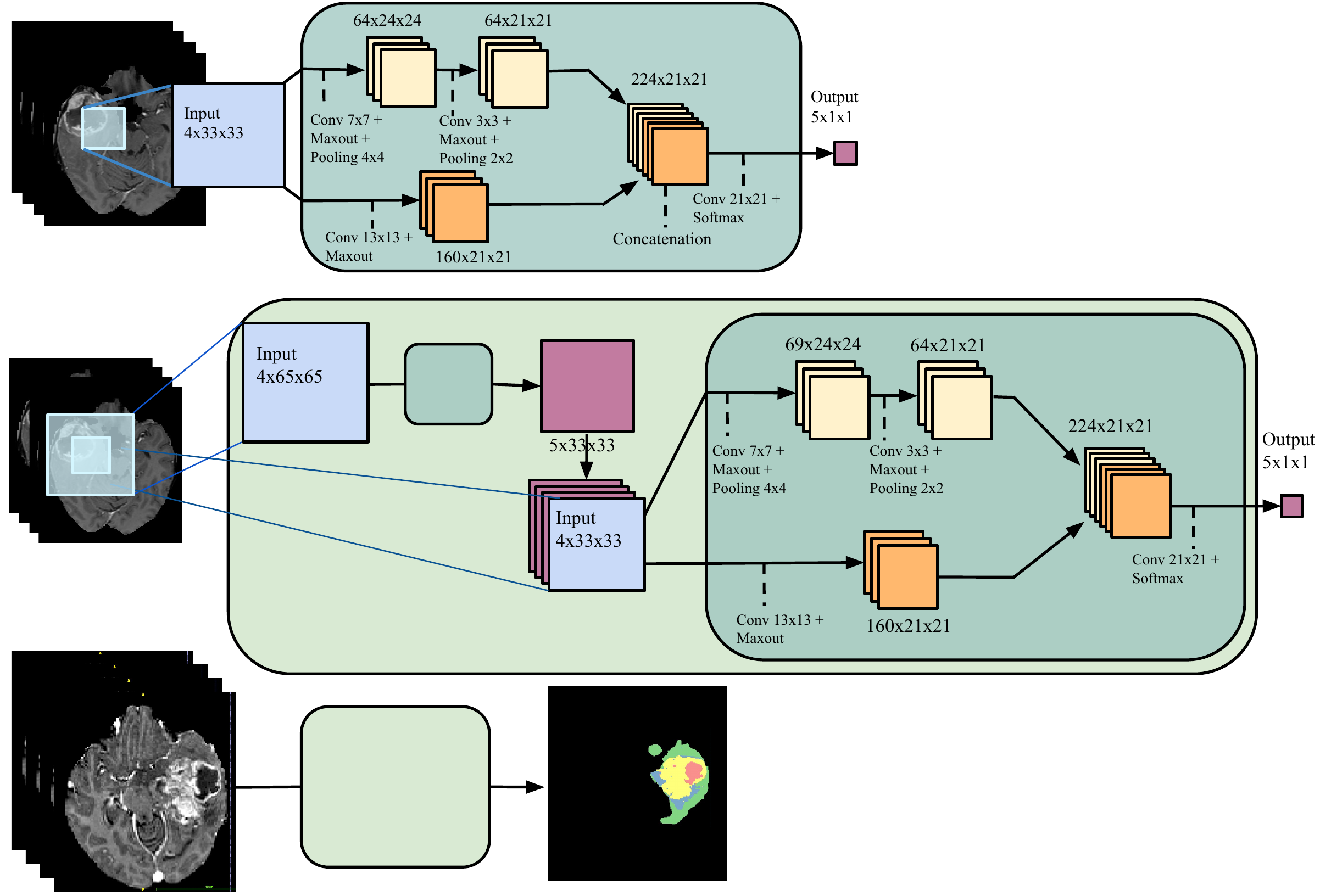}
\caption{The proposed architecture by Havaei et al.~\cite{Havaei2016}. First row: \textsc{TwoPathCNN}. The input patch goes through two convolutional networks each comprising of a local and a global path. The feature maps in the local and global paths are shown in yellow and orange respectively. Second row: \textsc{InputCascadeCNN}. The class probabilities generated by \textsc{TwoPathCNN} are concatenated to the input of a second CNN model. Third row: Full image prediction using \textsc{InputCascadeCNN}. }
\label{fig:havaei_model}
\end{figure}

 %The draw back of this approach is that there is no counter approach to deal with the stride convolution such as the ones presented in \cite{pinheiro2014} also predictions can be computationally expensive since the model is run for every overlapping patch. 
The TwoPathCNN by Havaei et al.~\cite{Havaei2016} consists of two pathways: a {\it local pathway} which concentrates on the pixel neighborhood information and a {\it global pathway} which captures more the global context of the slice. Their local path consists on 2 convolutional layers with kernel sizes of $7\times 7$ and $5\times 5$  respectively, while the global path consists of one convolutional layer with $11\times 11$ kernel size. In their architecture, they use {\it Maxout} \cite{Goodfellow_maxout_2013} as activation function for intermediate layers. Training patch size is set to $33 \times 33$, however during test time, the model is able to process a complete slice making the overall prediction time drop to a couple of seconds. This is achieved by implementing a convolutional equivalent of the dense layers. To preserve pixel density in the segmentation map, they use a stride of 1 in all max pooling and convolutional layers.\footnote{Using stride of $n$ means that every $n$ pixels will be mapped to $1$ pixel in the label map (assuming the model has one layer). This causes the model to loose pixel level accuracy if full image prediction is to be used at test time. One way to deal with this issue is presented by Pinheiro et al.~\cite{pinheiro2014}. Alternatively, we can use a stride of $1$ every where in the model.} This architecture is shown in Figure~\ref{fig:havaei_model}.

\begin{figure}[t!]
\centering
\includegraphics[width=1 \linewidth]{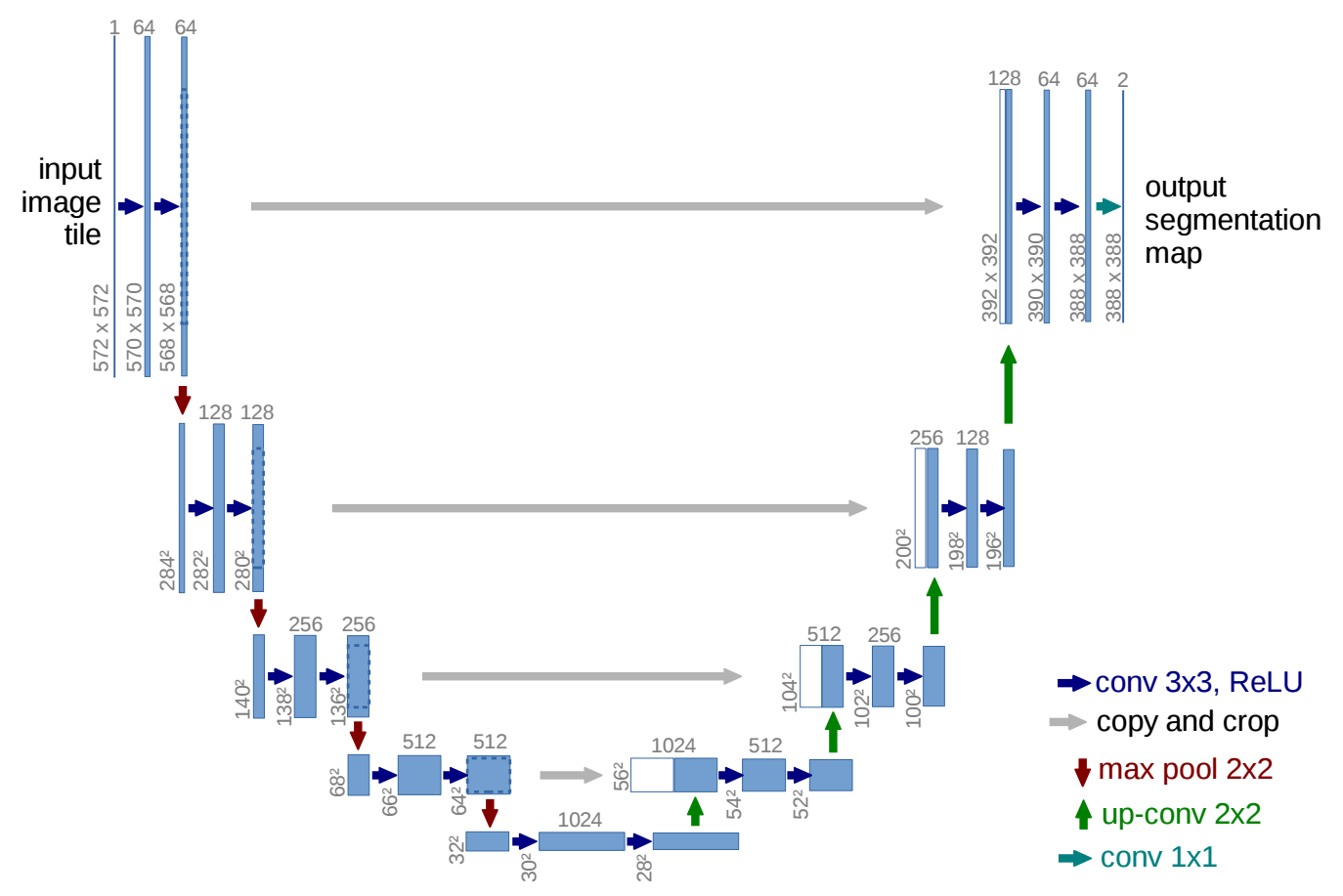}
\caption{U-Net: The proposed architecture by Ronneberger et al.~\cite{ronneberger2015u}.}
\label{fig:unet}
\end{figure}

\begin{figure}[t!]
\centering
\includegraphics[width=1 \linewidth]{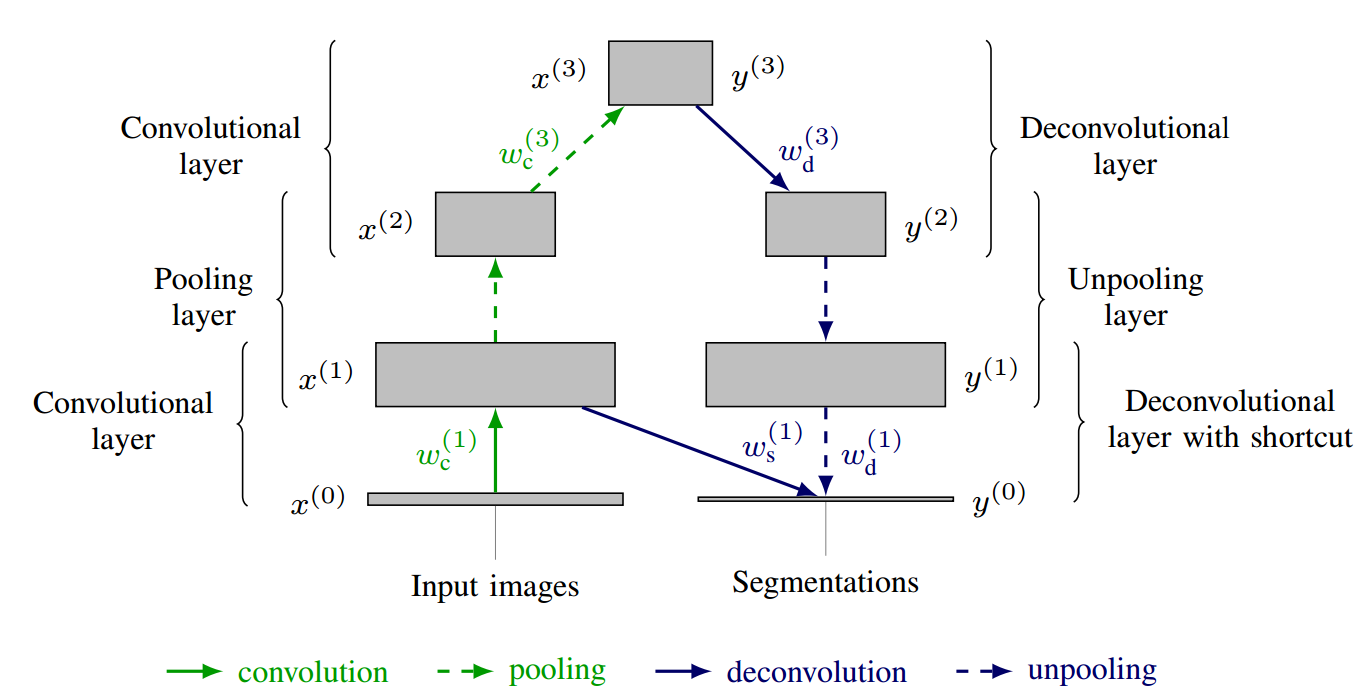}
\caption{CEN-s: The proposed architecture by Brosch et al.~\cite{brosch2016deep}.}
\label{fig:cen}
\end{figure}

%This technique has also been used by \cite{urban2014}. 
Havaei et al.~\cite{Havaei2016} also introduce a cascaded method where the class probabilities from a base model are concatenated with the input image modalities to train a secondary model similar in architecture to that of the base model. In their experiments, this approach refined the probability maps produced by the base model and brought them among the top 4 teams in BRATS 2015 \cite{havaeic}. 

Pereira et al.~\cite{pereira2015deep} also use a CNN with patch wise training and small kernel sizes (i.e. $3\times 3$) as suggested by \cite{simonyan2014}. This allowed them to have a deeper architecture while maintaining the same receptive field as shallow networks with larger kernels. They train separate models for HG and LG tumors. For the HG model, their architecture consists of 8 convolutional layers and 3 dense layers, while the LG model is a bit shallower, containing 4 convolutional layers and 3 dense layers. They use max pooling with a stride of 2 and dropout is used only on the dense layers. 
Leaky rectified linear units (LRLU)~\cite{maas2013rectifier} are used for the activation function of all intermediate layers. This method achieved good results in the BRATS 2015 challenge, ranking them among the top 4 winners. The authors also find {\it data augmentation} by rotation to be useful. That said, the method comes with a major inconvenience, which is for the user to manually decide the type of the tumor (LG or HG) to process. 

Dvorak et al.~\cite{dvorak2015structured} applied the idea of {\it local structure prediction} \cite{dollar2013structured} for brain tumor segmentation, where a dictionary of label patches is constructed by clustering the label patches into $n$ groups. The model is trained to assign an input patch to one of the $n$ groups. The goal is to force the model to take into account labels of the neighboring pixels in addition to the center pixel. 

The methods discussed above treat every MRI modality as a channel in the CNN. Rao et al.~\cite{rao2015} proposed instead to treat these modalities as inputs to separate convolutional streams. In this way, they train 4 separate CNN models each on a different modality. After training, these models are used as feature extractors where features from the last pooling layer of all 4 models are concatenated to train a random forest classifier. The CNNs share the same architecture of 2 convolutional layers of kernel size $5\times 5$ followed by 2 dense layers. Every CNN takes as input 3 patches of size $32 \times 32$, extracted from 3 dimensions (i.e. axial, sagittal, coronal) around the center pixel.

%\cite{hemis} proposed HeMIS, to deal with missing modalities. In this framework, a convolutional pathway is assigned to every present modality. The output feature maps of these {\it backend} models are fed to an {\it abstract} space where some statistics are derived. This abstract space acts as an input to a another CNN model (i.e. {\it frontend}) which performs segmentation. The whole model is trained end-to-end. This architecture is presented in Figure~\ref{fig:hemis}

 Segmentation problems in MRI are often 3D problems. However, employing CNNs on 3D data remains an open problem. This is due to the fact that MRI volumes are often anisotropic (especially for the FLAIR modality) and the volume resolution is not consistent across subjects. A solution is to pre-process the subjects to be isotropic \cite{Menze2014,guizard2015rotation}. However, these methods only interpolate the data and the result ends up being severely blurry when the data is highly anisotropic.  One way to incorporate information from 3D surroundings is to train on orthogonal patches extracted from axial, sagittal and coronal views. The objective would then be to predict the class label for the intersecting pixel. This is referred to as 2.5D in the literature \cite{rao2015,shin2016}.  Havaei et al.~\cite{Havaei2016} experimented with training on 2.5D patches. However, they argued that since BRATS 2013 train and test data have different voxel resolutions, the model did not generalize better than when only training on patches from the axial view. Vaidya et al.~\cite{vaidyalongitudinal} and Urban et al.~\cite{urban2014} used 3D convolutions for brain lesion and tumor segmentation. Using 3D convolution implies that the input to the model has an additional depth dimension. Although this has the  advantage of using the 3D context in the MRI, if the gap between slices across subjects varies a lot, the learnt features would not be robust. 
  In a similar line of thought, Klein et al.~\cite{KleinBatmanghelichWellsIII2015} also used 3D kernels for their convolutional layers, but with a different architecture. Their architecture consists of 4 convolutional layers with large kernel sizes on the first few layers (i.e. $12\times 12\times 12$, $7\times 7\times 7$, $5\times 5\times 5$, $3\times 3\times 3$) with input patch size of $41 \times 41 \times 41$. The convolutional layers are followed by 2 dense layers. 

Kamnitsas et al.~\cite{kamnitsas2015multi} used a combination of the methods above ~\cite{urban2014,Havaei2016,pereira2015deep}, applied to lesion segmentation. In their 11 layer fully convolutional network which consisted of 2 pathways similar to \cite{Havaei2016}, they used 3D convolutions with small kernel sizes of $3\times 3\times 3$. Using this model, they ranked among the winners of the ISLES 2015 challenge. 
  
Stollenga et al.~\cite{stollenga2015parallel} used a long short-term memories (LSTM) network applied to 2.5D patches for brain segmentation.  

As opposed to methods which use deep learning in a CNN framework, Vaidhya et al.~\cite{vaidhya2015multi} used a multi-layer perceptron consisting of 4 dense layers. All feature layers (i.e. the first 3) were pre-trained using denoising auto-encoder as in \cite{vincent2010stacked}. The input consists of 3D patches of size $9\times 9\times 9$. Training is performed on a resampled version of the BRATS dataset, which balances the number of class patches. However, similar to \cite{Havaei2016}, fine-tuning is done on the original dataset with imbalanced classes to reflect the real distribution of label classes.

Inspired by \cite{masci2011stacked}, Brosch et al.~\cite{brosch2015} presented the convolutional encoder networks (CEN) for MS lesion segmentation. The model consists of 2 parts; the encoder part which decreases the resolution of the feature maps and the up sampling part (also known as the decoder part) which increases the resolution of the feature maps and performs pixel level classification \footnote{In the literature this way of up sampling is some times wrongly referred to as {\it deconvolution}.}. The encoder consists of 2, 3D convolutional layers in {\it valid mode}\footnote{Valid mode is when kernel and input have complete overlap.} with kernel size $9\times 9\times 9$ in both layers, followed by an ReLU activation function.
The up sampling part of the model consists of convolutions in {\it full mode}\footnote{Full mode is when minimum overlap is a sufficient condition for applying convolution.} which results in up sampling the model. Balancing label classes is done by introducing weights per class in the loss function. %This is done by using sensitivity and specificity in the loss function which takes into account a weight for classes [TODO fix this sentence to make sense]
They improved on this method in \cite{brosch2016deep} by introducing CEN-s, where they combine feature maps from the first hidden layer to the last hidden layer. As shown in Figure.~\ref{fig:cen} and Figure~\ref{fig:unet}, this model is very similar to the U-Net by Ronneberger et al.~\cite{ronneberger2015u} with a difference in the way the up sampling step is applied. While U-Net uses interpolation for up sampling, CEN-s uses convolutions and the transformation weights are learnt during training. Also U-Net is deeper with 11 layers, while CEN-s contains only 4 layers. Inspired by ~\cite{resnet}, Drozdzal et al.~\cite{resunet} expand U-Net by adding short skip connections which allows them to train very deep models.

Combining feature maps from shallow layers to higher layers (also referred to as {\it skip} or {\it shortcut} connections) are popular in semantic segmentation \cite{long2015fully,hariharan2015hypercolumns}.

%A straight forward way is to apply CNN on 2D slices and process the MRI slice by slice [deeporgan][unet][havaei]\cite{vivantiautomatic}.  
%\cite{dou2016} 

\begin{figure}[tp]
\centering
\includegraphics[width=1 \linewidth]{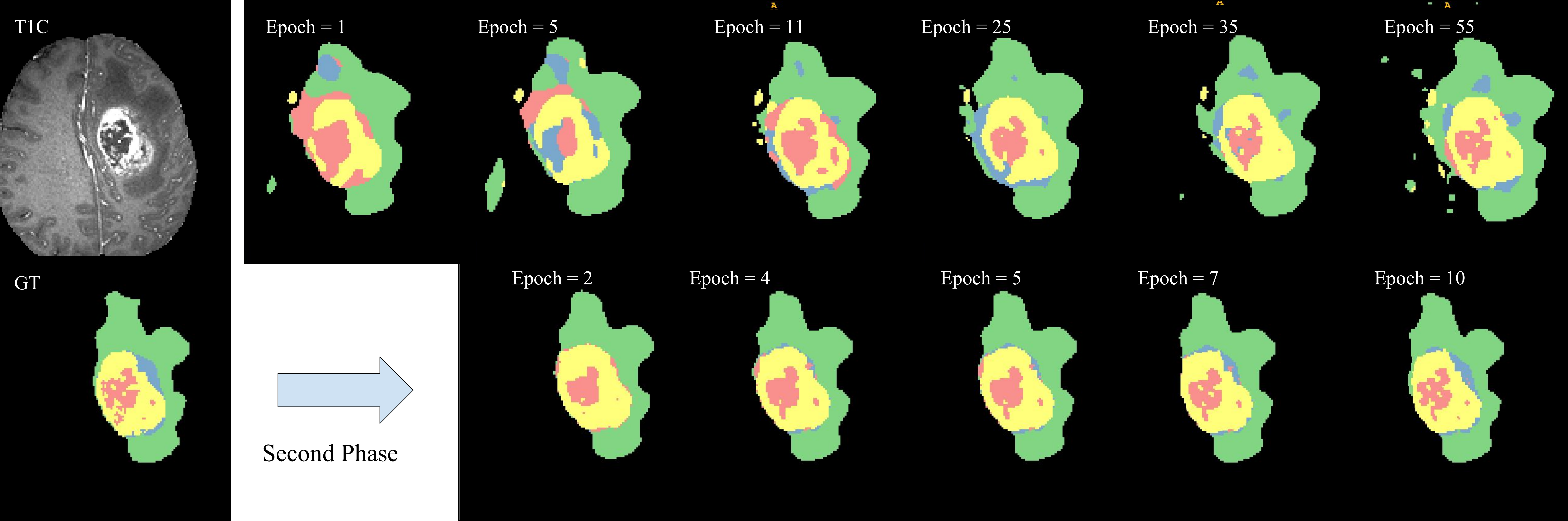}
\caption{Effect of second phase training proposed by \cite{Havaei2016}. The figure shows how the second phase regularizes the predictions and removes false positives.
}
\label{fig:scheduled_prediction_0}
\end{figure}

\section{Open Problems}
\label{open-problems}
\subsection{Preparing the dataset}
Preparing the dataset in a proper way can play a key role in learning. In this chapter, we discuss important aspects of dataset preparation for medical imaging. 

\subsubsection{Pre-processing}
As mentioned before, the grayscale distribution of MR images depends on the acquisition protocol and the hardware. This makes learning difficult since we expect to have the same data distribution from one subject to another. %This can be overcome when having access to infinite amount of data which is not the case in medical imaging. 
Therefore, pre-processing to bring all subjects to similar distributions is an important step.  Also, it is desirable that all input modalities have the same intensity range, so one modality does not have prior advantage over others in deciding the output of the model. Among the many pre-processing approaches reported in the literature, the following are the most popular: 
\begin{itemize}
    \item Applying the N4/N3 bias field correction \cite{tustison2015,Havaei2016,gotz2014extremely,zikic2014,Kwon2014,guizard2015rotation,dvorak2015structured}. Kleesiek et al.~\cite{Kleesiek2014} and Urban et al.~\cite{urban2014} did not apply bias field correction, instead, they performed intensity normalization with mean CSF value, which they claim to be more robust and effective.
    \item Truncating the 1\% or 0.1\% quantiles of the histogram to remove outliers from all modalities \cite{tustison2015,Havaei2016,vaidhya2015multi}.
    \item Histogram normalization, which is mostly done by matching the histogram of every modality to their corresponding template histogram. \cite{bakas2015segmentation,pereira2015deep,vaidhya2015multi,guizard2015rotation}.
    \item Normalizing modalities \cite{Havaei2016,dvorak2015structured} or the selected training patches \cite{pereira2015deep} to have zero mean and unit variance. 
\end{itemize}

\subsubsection{Shuffling}
Introducing the data to the model in a sequential order results in biasing the gradients and can lead to poor convergence. By sequential order, we mean training first on data (i.e. patches or slices) extracted from a subject, then training on data extracted from another subject, and so on until the end of the training set. Depending on the dataset, MRI subjects can be very different in terms of noise and even intensity distribution. Therefore, it is important to shuffle the entire dataset so the model does not overfit to the current training subject and forget its previous findings. It is desirable that the distribution from which we introduce training examples to the model does not change significantly.
An advantage of patch wise training over full image training is that patch wise training allows us to fully shuffle the dataset. This means, in patch wise training, every mini batch contains patches from different slices of different subjects while in full image training, there is no shuffling at pixel level.

\subsubsection{Balancing the dataset}
A dataset is imbalanced when class labels are not approximately equally represented. %The issue of class imbalance becomes noticeable when applying machine learning algorithms to real-world applications. 
Unfortunately, brain imaging data are rarely balanced due to the small size of the lesion compared to the rest of the brain.  For example, the volume of a stroke is rarely more than $1\%$ of the entire brain and a tumor (even large glioblastomas) never occupy more than $4\%$ of the brain.  Training a deep network with imbalanced data often leads to very low, true positive rate since the system gets biased towards the one class that is over represented.  

Ideally, we want to learn features invariant to the class distribution. This can be done through balancing the classes in the dataset.  One approach is to take samples from the training set so we get an equal number of samples for every class. Another approach is to weight the loss for the training examples from different classes based on the frequency of appearance of every class in the training data \cite{ronneberger2015u} \cite{brosch2015}.
Sampling from the training set can be done randomly \cite{roth2015,roth2015deep,roth2014new}, or follow an importance sampling criterion to help the model learn features we care about (for example border between classes).
In Havaei et al.'s~\cite{Havaei2016} patch wise training method, the importance sampling is done by computing the class entropy for every pixel in the ground truth and giving training priority to patches with higher entropy. In other words, patches with higher entropy, contain more classes which makes them good candidates to learn the border regions from. 

Training on a balanced dataset makes the model believe all classes are equiprobable and thus may cause some false positives.  In order to compensate for this, one can account for the imbalanced nature of the data with a second training phase, during which, only the classification layer is trained and other feature layers are fixed. This allows to regularize the model and remove some false positives. The effect of the second phase training is presented in Fig~\ref{fig:scheduled_prediction_0}.
Ronneberger et al.~\cite{ronneberger2015u} took a different approach which is best suited for full image training. In their approach, they compute the distance of every pixel to class borders and, based on that, a weight is assigned to every pixel. A weight map is created for every training image and is used in the loss function to weight every sample differently. 

Pereira et al.~\cite{pereira2015deep} balance classes mainly by data augmentation. In their case, data augmentation can be either a transformation applied on a patch or simply using patches from similar datasets. For example, using patches from brains with high-grade glioma when training a low-grade glioma  model.

\subsection{Global information}
Adding context information has always been a subject of interest in medical image analysis \cite{ali2015multi,corso2006multilevel,corso2008efficient}.  Since anatomical regions in closeup view can appear similar and borders may be diffused in some parts due to lack of contrast or other artifacts, additional context is needed to localize a region of interest. 

In a CNN, it is possible to encode more contextual information by increasing the portion of the input image that each neuron sees (directly or indirectly). Although it is possible to increase the receptive filed of a neuron on the input image through series of  convolutional and pooling layers of stride $1$, using strides greater than $1$ is computationally more efficient and results in more robust features. By doing so, the model looses precision of spatial information which is needed for segmentation purposes. 
To take advantage of both worlds (i.e. having spatial precision while learning robust features through pooling layers) encoder-decoder type architectures can be used. Ronneberger et al.~\cite{ronneberger2015u} and Brosch et al.~\cite{brosch2016deep} learn a global understanding of the input by down sampling the image (through series of convolutional and pooling layers) to smaller size feature maps. These feature maps are later up sampled in the decoder section of the model and combined with feature maps of lower layers that preserve the spatial information (see Figure~\ref{fig:unet} and Figure~\ref{fig:cen}). 

Havaei et al.~\cite{Havaei2016} take a different approach where feature maps from $2$ convolutional streams (using the same input) are concatenated before going through the classification layer. This two pathway approach, allows the model to learn simultaneously local and global contextual features (see Figure~\ref{fig:havaei_model}).

\subsection{Structured prediction}
Although CNNs provide powerful tools for segmentation, they do not model spatial dependencies in the segmentation space directly. To address this issue, many methods have been proposed to take the information of the neighboring pixels in the label image into account. 
These methods can be divided into two main categories. The first category are methods which consider the information of the neighboring labels in an {\it implicit} way, while providing no specific pairwise term in the loss function.  An example of such an approach is provided by Havaei et al.~\cite{Havaei2016} which refine predictions made by a first CNN model by providing the posterior probabilities over classes as extra inputs to a second CNN model. Roth et al.~\cite{roth2015} also use a cascaded architecture to concatenate the probabilities of their first convolutional model with features extracted from multiple scales in a {\it zoom out} fashion \cite{mostajabi2015feedforward}.  %If weights re shared between the models then the whole model acts as a recurrent neural network. 
The second category of methods are ones that {\it explicitly} define a pairwise term in the loss function which is usually referred to as Conditional Random Field (CRF) in the literature. Although it is possible to train the CNN and CRF end to end, usually for simplicity, the CRF is trained or applied as a post processing secondary model to smooth the predicted labels. The weights for the pairwise terms in the CRF can be fixed \cite{havaei2014}, determined by the input image \cite{havaei2014} or learned from the training data~\cite{roth2015}. In their work, Roth et al.~\cite{roth2015} trained an additional CNN model between pairs of neighboring pixels. 

Post-processing methods based on {\it connected components} have also proved to be effective to remove small false positive blobs \cite{vaidhya2015multi,Havaei2016,pereira2015deep}.  In \cite{roth2015}, the authors also try 3D isotropic Gaussian smoothing to propagate 2D predictions to 3D and according to them, Gaussian smoothing was more beneficial than using a CRF.

%\subsection{Multi-resolution}
%in \cite{roth2015} patches at different scales are introduced to the model as different input channels. . 

\subsection{Training on small or incomplete datasets}
Deep neural networks generalize better on new data if a large training set is available. This is due to the large number of parameters present in these models. However, constructing a medical imaging dataset is an expensive and tedious task which causes datasets to be small and models trained on these datasets prone to overfitting. Even the largest datasets in this field does not exceed a few hundred subjects. This is much smaller than datasets like ImageNet, which contains millions of images. 

Another problem arises from incomplete datasets. Medical imaging datasets are often multi-modal with images from MRI acquisitions (T1, T2, PD, DWI, etc.) \cite{Menze2014,li2014}. However, not all modalities are always available for every subject. How to effectively use the incomplete data rather than simply discarding them is an open question. Another scenario is how to generalize on subjects with missing modalities.  
In this section we review several effective approaches to train on small and/or incomplete datasets 

\subsubsection{Data augmentation}
Increasing the size of the dataset by data augmentation is commonly employed in machine learning to enrich a dataset and reduce overfitting \cite{Krizhevsky-2012-small}. Flipping the image, applying small rotations and warping the image are common practices for this purpose \cite{Krizhevsky-2012-small,cirecsan2013,ronneberger2015u}. Roth et al.~\cite{roth2015} and Ronneberger et al.~\cite{ronneberger2015u}  use non-rigid deformation transformations to increase the size of their datasets and report it to be a key element in achieving good results. 
The type of data augmentation technique depends on the anatomy of the data and the model being used. For example, Pereira et al.~\cite{pereira2015deep} only tested with rotation for data augmentation because the label of the patch is determined by the center pixel and so warping or applying translations might change the position of the center pixel. They used angles multiple of $90^{\circ}$ and managed to increase the size of the dataset 4 times. They found data augmentation to be very effective in their experiments. 
%In general, Medical imaging datasets are not big enough for deep learning and so data augmentation can be  applied to increase the size of the dataset. 

\subsubsection{Transfer learning}
Deep learning has made significant breakthroughs in computer vision tasks due to training on very large datasets such as ImageNet. ImageNet contains more than 1.2 million training examples on over 1000 classes. To improve generalization on smaller datasets, it is common to first train a {\it base} model on a large dataset such as ImageNet and then fine-tune the learnt features on a second {\it target} model which is often much smaller in size. Yosinski et al.~\cite{yosinski2014}  showed that the transferability of the features depends on how general those features are. The transferability gap increases as the distance between the tasks increase and also as the features become less general in higher levels. However, initializing weights from a  pre-trained model (preferably on a large dataset), is still better than initializing weights randomly.  

Transfer learning can take different forms. One way is to generate features from the base model and then use those features to train a classifier such as SVM or logistic regression \cite{bar2015,van2015off,arevalo2015convolutional}.
Bar et al.~\cite{bar2015} used an ImageNet pre-trained base model to extract features. These features are concatenated with other hand-crafted features before being introduced to an SVM classifier. % used for chest pathology
 Van et al.~\cite{van2015off} used {\it overfeat} pre-trained weights to generate features for lung tumor detection. To address the overfeats $3$ input channels, $3$ 2d patches are extracted from axial, saggital and coronal views. SVM is used as classifier. 

Although this way of transfer learning has proved to be somewhat successful, the degree of its usefulness depends on how similar the source and target datasets are. If not very similar, a better alternative is to fine-tune the features on the target dataset~\cite{chen2015standard,carneiro2015,gaoholistic,margeta2015fine}. 
Gao et al.~\cite{gaoholistic} used this fine-tuning scheme to detect lung disease in CT images. To account for the 3 color channels of the base model which has been pre-trained on ImageNet, 3 attenuation scales with respect to lung abnormality patterns are captured by rescaling the original 1-channel CT image.  
Carneiro et al.~\cite{carneiro2015} uses this method to reach state-of-the-art results on the InBreast dataset.
Shin et al.~\cite{shin2016} reported experimental results in 3 transfer learning scenarios for Lymph node detection. 1) No transfer learning 2) transfering the weights from a base model and only training the classification layer (i.e. weights from other layers are frozen), 3) transfering the weights from a base model and fine-tuning all layers. According to their experiments, the best performance was achieved in the 3rd scenario where the weights of the target model are initialized from the weights of a previously trained base model and then all layers are fine-tuned on the Lymph node dataset. Also, scenario 1 achieved the worst performance. This is expected since the two datasets are very different and the features learnt by a model trained on ImageNet are not general enough to be used as is on a medical imaging dataset. Tajbakhsh et al.~\cite{tajbakhsh2016convolutional} conducted a similar study on transferring pre-trained weights from {\it AlexNet} trained on ImageNet to 4 medical imaging datasets. Based on their findings, initializing the weights to a pre-trained model and fine-tuning all layers should be preferred to training from scratch, regardless of the size of the dataset. However, if the target dataset is smaller we should be expecting a better gain in performance compared to when the target dataset is sufficiently large. They also observed that transfer learning increases the convergence speed on the target model. Also, since the natural scene image datasets such as ImageNet are very different from medical imaging datasets, we are better off fine-tuning all the layers of the model as opposed to only fine-tuning the last few layers. Van et al.~\cite{van2015off} also came to a similar conclusion.

Another approach to transfer learning is to initialize the model to weights which have been pre-trained separately in an unsupervised way using models such as {\it Autoencoders} or {\it RBMs} \cite{larochelle2007}. This allows the weights of the target model to be initialized in a better {\it basin of attraction} \cite{erhan2010}. 
In their lung segmentation problem where they had access to a large un-annotated dataset and a smaller annotated dataset, Schlegl et al.~\cite{schlegl2014unsupervised} used convolutional restricted boltzmann machine to pre-train a CNN model in an unsupervised fashion. A shallow model is used as it helps the unsupervised model to learn more general features and less domain specific features.

\subsubsection{Missing modalities}
Different modalities in MRI need to be acquired separately and it often happens that different subjects are missing some modalities. The most common practice is to prepare the dataset using modalities that exist for most subjects. This leads to either discarding some subjects from the dataset or discarding some modalities which are not present in all subjects.
Another approach is to impute the missing modalities by zero or the mean value of the missing modality.
Li et al.~\cite{li2014} used a 3 dimensional CNN architecture to predict a PET modality given a set of MRI modalities. 
Van et al.~\cite{van2015} proposed to synthesize one missing modality by sampling from the hidden layer representations of a Restricted Boltzmann Machine (RBM). They perform their experiments on BRATS 2013 using a patch wise training approach. For every training patch, they train the RBM with every modality to learn the joint probability distribution of the four modalities. At test time, when only one of the modalities is missing, they can estimate the missing modality by sampling from the hidden representation vector. 

\cite{hemis} proposed HeMIS, a system for dealing with missing modalities. Contrary to other approaches, HeMIS does not require all modalities to be preset. Every modality gives a vote to an {\it abstract} space, the more modalities present, the stronger statistics are derived and the better performances get.  However, HeMIS is not dependant on the presence of the least informative modalities.  As a result, the performance of the model drops gracefully. %For example even if T1c modality is not present, the model can still get high accuracy on the complete tumor region (see Figure~\ref{fig:hemis_result}).

\section{Future Outlook}
Although deep learning methods have proven to have potential in medical image analysis applications, their performance depends highly on the quality of the pre-processing and/or the post processing. These methods tend to perform poorly when input data do not follow a common distribution which is often the case. Learning robust representations which are invariant to the noise introduced by the acquisition is needed. Unsupervised learning or weakly supervised learning might hold the key to this problem. Also methods based on domain adaptation might help us learn representations which can better generalize across datasets.  

\small{
\bibliographystyle{splncs03}

}

\end{document}